\documentclass[letterpaper, 10 pt, conference]{template-files/ieeeconf/ieeeconf}

\IEEEoverridecommandlockouts
\usepackage{graphicx}
\usepackage{amsmath}
\usepackage{amssymb}
\usepackage{cite}
\usepackage{booktabs}
\usepackage{multirow}
\usepackage{xcolor}
\usepackage{colortbl}
\usepackage[T1]{fontenc}
\usepackage{lmodern}
\let\labelindent\relax
\usepackage{enumitem}

\graphicspath{{./figures/}}
\title{\LARGE \bf
Fast and Reliable Open-Vocabulary Object Search with Absence Detection from Robot Memory
}

\author{Nicholas Paul$^{1}$, Devson Butani$^{1}$, and Eric Martinson$^{1}$%
\thanks{$^{1}$Lawrence Technological University, Southfield, MI, USA.
{\tt\small npaul@ltu.edu, dbutani@ltu.edu, emartinson@ltu.edu}}}

\begin{document}

\maketitle
\thispagestyle{empty}
\pagestyle{empty}

%%%%%%%%%%%%%%%%%%%%%%%%%%%%%%%%%%%%%%%%%%%%%%%%%%%%%%%%%%%%%%%%%%%%%%%%%%%%%%%%
\begin{abstract}

Mobile manipulators deployed in everyday human environments, such as homes for assistive care or flexible automation, often need to locate specific objects on demand from accumulated observations.
Equally important, they must reliably determine when a target is absent before committing to navigation or interaction.
Existing approaches demand either slow exhaustive re-processing or expensive 3D representations, and are evaluated on recall and accuracy rather than specificity, leaving their ability to abstain on missing targets unmeasured.
We find existing systems that search using text or visual embedding almost never abstain on absent objects achieving at most 9\% specificity.
To support accurate search at speed, we present a two-stage architecture that decouples fast candidate retrieval from verified selection, enabling both high recall and reliable absence detection from sequential RGB observations with pose.
We evaluate eight retrieval-selection configurations plus an exhaustive search baseline across ScanNet apartment scenes, including significant numbers of absent objects.
Adding a visual verification step to caption-based retrieval matches or exceeds exhaustive search reliability while requiring an order of magnitude fewer detector calls.
We reproduce ScanNet scores on a physical robot operating entirely on-board with a single consumer GPU running in real time.

\end{abstract}

%%%%%%%%%%%%%%%%%%%%%%%%%%%%%%%%%%%%%%%%%%%%%%%%%%%%%%%%%%%%%%%%%%%%%%%%%%%%%%%%
\section{INTRODUCTION}

Advances in robotic technologies and artificial intelligence are rapidly transforming the adoption of mobile manipulators and humanoid robots in contexts such as healthcare, in-home assistance, and industry\cite{tong2024advancements}. A home assistant asked to fetch medication, a warehouse robot verifying inventory, or a search-and-rescue robot platform looking for information all face the same fundamental problem: the robot has accumulated observations over time and must now answer "where is X?" or equally important, "X is not here." Getting either of these questions wrong erodes a user's trust and the perception of reliability\cite{radka2025attitudes}.

The simplest solution to reliable object retrieval from a pre-recorded dataset is exhaustive search -- reprocessing the entire dataset with a detection algorithm like SAM3~\cite{sam3} and identifying the pose of the robot associated with the Top 1 image. However this comes at a significant compute cost: if SAM3 runs at 3 Hz on the platform's local compute, then it will take >5 minutes to search a 1000 image dataset for the new object; which is an interminable delay in many scenarios. Alternative solutions pre-process the dataset, constructing fused 3D representations of potential object locations. Approaches such as ConceptGraphs~\cite{gu2024conceptgraphs} and HOV-SG~\cite{werby2024hierarchical}, are powerful but rely on fully reconstructed scenes or RGB-D information which is computationally expensive and not always available. Hybrid approaches that pair 3D representations with exhaustive search improve performance at the cost of even greater computational overhead~\cite{martinson2025augmenting}. Other approaches such as ReMEmbR~\cite{anwar2025remembr} and Embodied-RAG~\cite{xie2024embodied} are lightweight and avoid 3D reconstruction, but focus primarily on retrieval effectiveness without formally evaluating the ability to reject.

To retain efficiency but operate at speed, we introduce a two-stage architecture that explicitly separates retrieval from selection, allowing broad retrieval with high recall from strict selection of a final result.
Visual verification in the second stage gives the system a mechanism to abstain.

We present a modular two-stage architecture that enables reliable absence detection from sequential RGB observations without 3D reconstruction. A systematic evaluation of eight retrieval-selection combinations plus an exhaustive baseline, demonstrates that the selection stage raises specificity from 9\% to as high as 92.5\% (Caption/SAM3) while requiring only a fraction of detector calls. Our system, which requires less than 2 seconds per query on local compute, matches or exceeds the exhaustive baseline which runs at approximately 5 minutes per query on a 1000-image dataset on the same compute. The generality of this system is further validated on a physical platform running entirely on local consumer-grade compute.

%%%%%%%%%%%%%%%%%%%%%%%%%%%%%%%%%%%%%%%%%%%%%%%%%%%%%%%%%%%%%%%%%%%%%%%%%%%%%%%%
\section{RELATED WORK}

Language-grounded scene representations enable robots to query their environment using natural language. ConceptGraphs~\cite{gu2024conceptgraphs} builds an object-centric graph with explicit relations and HOV-SG~\cite{werby2024hierarchical} uses a hierarchical approach which excels at breaking down navigation tasks. VLMaps~\cite{huang2023visual} overlay language features onto an existing 3D map. Other semantic representations use fields instead of graphs. CLIP-Fields~\cite{shafiullah2022clip} creates a 3D reconstruction represented as a feature field. LERF~\cite{kerr2023lerf} leverages a NeRF trained from RGB-D images and poses and augments it with language features. A common limitation among these representations is their heavy reliance on RGB-D data, point clouds, or pre-trained neural models with limited flexibility.

To bypass the overhead of 3D reconstruction, many methods have pivoted toward retrieval-augmented memory. For instance, Embodied-RAG~\cite{xie2024embodied} stores information as a hierarchical forest atop a topological map to enable multi-level queries. Most similar to our approach is ReMEmbR~\cite{anwar2025remembr}, which stores VLM-generated captions from short video clips in a vector database for fast retrieval. Despite their efficiency, these retrieval-augmented frameworks primarily prioritize recall and QA quality over precision and principled rejection. When a queried object is absent, these systems often struggle to abstain, instead attempting to answer based on weak or irrelevant retrieved memories. ReMEmbR's query loop, for instance, is explicitly designed to always provide an answer with no ability to reject. The authors illustrate this consequence: a query for a \textit{soda machine} that was not present returned a confident, but incorrect, match rather than abstaining. Even pipelines like OK-Robot~\cite{liu2024ok}, which uses a two-stage approach of CLIP-based retrieval followed by LangSam verification, identify incorrect semantic retrieval as a leading failure mode.

Open-vocabulary detectors like Grounding-DINO~\cite{liu2024grounding}, OWLv2~\cite{minderer2023scaling}, YOLO-World~\cite{cheng2024yoloworld}, and SAM3~\cite{sam3} all support answering questions such as "find X in this image" using plain text queries. Grounding DINO is a strong vision-language backbone widely used as the detection front-end in open-vocabulary pipelines. Grounded-SAM~\cite{ren2024grounded} expands on this by providing segmentation of an image. OWLv2 and YOLO-World also use open-vocabulary to score and localize candidate regions, but like the others, these are not designed to reason about non-existence. SAM3, by contrast, introduces a dedicated presence head that first estimates whether a queried concept exists before localizing it, reducing false positives and negatives compared to previous approaches. Our architecture builds upon this capability by integrating memory retrieval to determine if an object exists in \textit{any} observation, requiring only RGB data and image poses. Additionally, our evaluation focuses on the ability to reject absent objects, an important metric not formally evaluated in these existing methods.

%%%%%%%%%%%%%%%%%%%%%%%%%%%%%%%%%%%%%%%%%%%%%%%%%%%%%%%%%%%%%%%%%%%%%%%%%%%%%%%%
\section{APPROACH}

\begin{figure}[t]
    \centering
    \includegraphics[width=0.95\columnwidth]{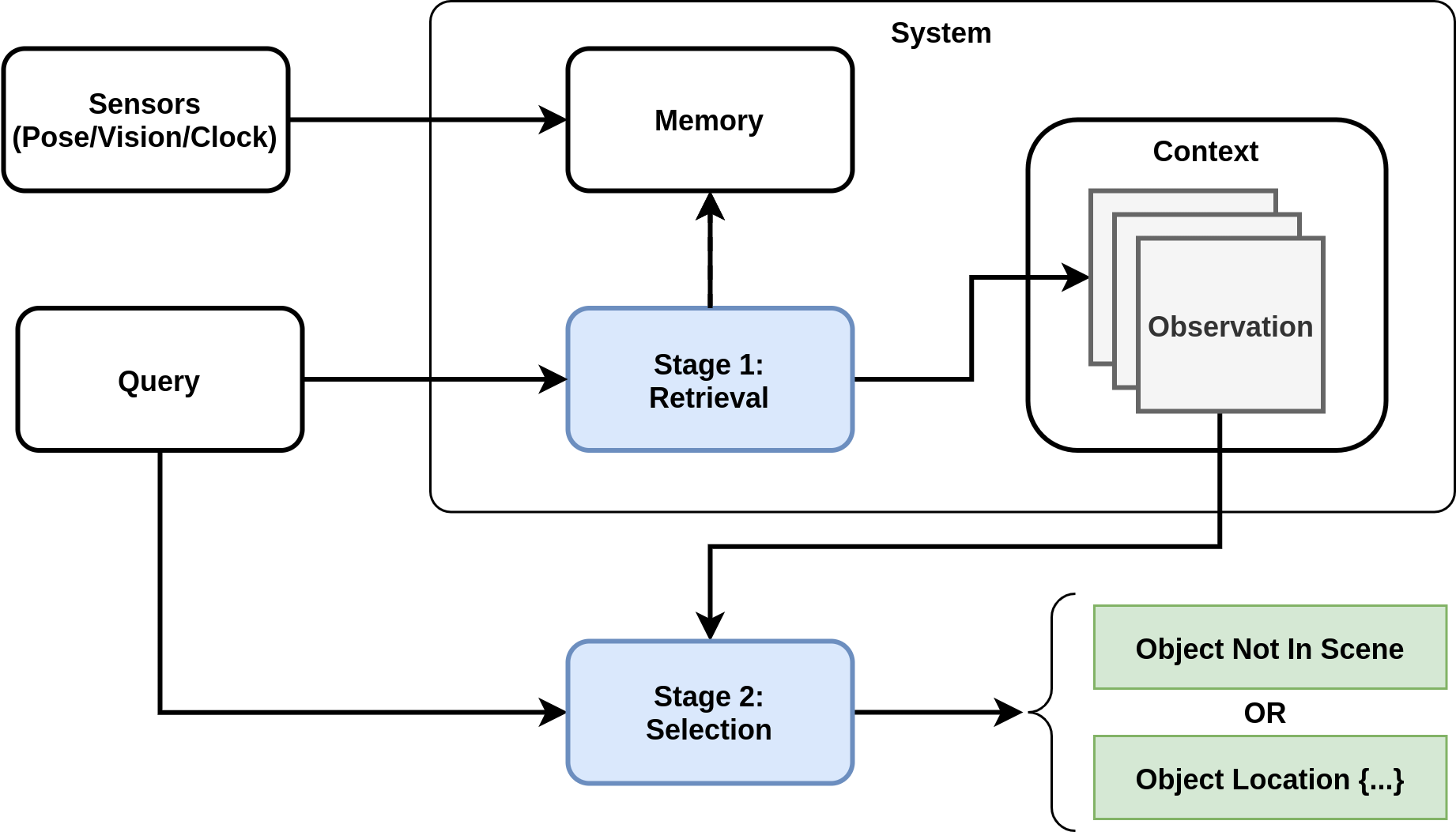}
    \caption{System overview. Observations are indexed in a vector DB. Stage~1 retrieves the top-$K$ candidates; Stage~2 selects a final answer or returns "not found."}
    \label{fig:arch}
\end{figure}

Our system operates on a store of observations collected from the same environment. Each observation consists of a camera frame, an image pose, a timestamp, and an embedding (either a CLIP visual embedding or a text embedding of a VLM caption). The image pose can be generated with SLAM architectures or post-hoc with image registration models such as COLMAP~\cite{schonberger2016structure}. Given a natural language query $q$, the system must return the observation best matching $q$ or \textbf{absent} (Fig.~\ref{fig:arch}). The pipeline proceeds in two stages: Stage 1 retrieves the top-$K$ candidate observations from the full store using fast similarity search and Stage 2 applies one of several selection methods to make the final determination of which observation to select or reject the query.

\subsection{Stage 1: Candidate Retrieval}

The first stage of the architecture serves as a filter, narrowing the full observation set down to the top-$K$ candidates to be processed by the subsequent verified selection stage. Two retrieval strategies are currently being evaluated: visual or caption-based.

With \textbf{visual} retrieval, CLIP~\cite{radford2021learning} (using ViT-B/32) produces a 512-dimensional embedding vector for each observation frame. At query time, $q$ is embedded using CLIP's text encoder and compared against all visual embeddings via cosine similarity, returning the $K$ nearest.

For \textbf{caption based} retrieval, a text description of each observation is generated using Gemma 3~\cite{gemma3_tech_report} (12B-parameter vision-language model). The resulting captions are then chunked and embedded with Nomic Embed Text~\cite{nussbaum2024nomic}. At query time, the same text embedding model encodes the query, and the nearest $K$ chunks are retrieved using cosine similarity. While only the relevant chunk is embedded and retrieved, the full caption text is returned as the observation. 

All embeddings, both visual and caption-based, are stored and queried using the vector database Milvus~\cite{wang2021milvus}. While caption-based retrieval is slower due to the overhead of the initial VLM inferences, it captures richer semantic content that CLIP embeddings do not, producing more diverse candidates drawn from different parts of the scene rather than temporally adjacent frames that share similar visual appearance.

\subsection{Stage 2: Verified Selection}

The goal of the selection stage is to select the single best observation from the set of top-$K$ identified in the retrieval phase or, crucially, reject all if not present. We evaluate 4 selection methods.

% \begin{itemize}
\begin{enumerate}[leftmargin=15pt]
    \item \textbf{Top-1} receives the cosine similarity score for the top-$K$ results and always returns the highest-ranked result. It is therefore not capable of rejecting and used as a baseline.
    \item \textbf{SAM3} runs SAM3~\cite{sam3} on each image in the candidate set. The image with the highest score value is selected. If no image returns a positive result from SAM3, the query is considered absent.
    \item \textbf{LLM} receives the full caption of each image and is prompted to select a single candidate or reject if it believes the object is not present with a structured response. The LLM is free to infer whether or not it thinks an object is present even if it is not explicitly mentioned in the caption. For instance if the query is "dish soap" and captions of retrieved observations mention a kitchen sink and a bottle of cleaner, an LLM may infer that the likelihood of "dish soap" present in the observation is high and respond with a positive result even if it is not explicitly mentioned. This improves accuracy since captions cannot explicitly mention every possible query that could be relevant to the image. On the other hand, this increases the rate at which an LLM may guess incorrectly and return a false detection when an object is indeed not present.
    \item \textbf{LLM+SAM3} is the same as \textit{LLM}, except the LLM receives both the caption and the SAM3 detection scores for the query in its context. For this method, we first run SAM3 on all candidates. Along with the caption we include either the SAM3 score for the query, or mention that the image verification did not identify any objects. The LLM is prompted in a similar way to the previous selection method and ultimately makes the final decision of which observation to select or to reject. This means the LLM may decide to select an observation even if there are no objects according to SAM3 or it may decide to abstain even if SAM3 produces a positive result.
% \end{itemize}
\end{enumerate}

If the object is successfully retrieved, all algorithms return the single best image matching the target query and the associated capture pose that a robot can use to find the object again.
% How often does the LLM select a different image than the highest SAM3 score?? How often do they disagree?
The complete set of retrieval selection configurations evaluated are summarized in Table~\ref{tab:combinations}. Because standard recall and $F_1$ reward a system that always answers, our evaluation reports results primarily with $F_{1s}$, a specificity-weighted score that rewards correct abstention as well as correct retrieval.

%%%%%%%%%%%%%%%%%%%%%%%%%%%%%%%%%%%%%%%%%%%%%%%%%%%%%%%%%%%%%%%%%%%%%%%%%%%%%%%%
\section{SCANNET EVALUATION}

\subsection{Dataset and Query Design}

We evaluate all 9 methods on 8 ScanNet~\cite{dai2017scannet} apartment scenes with 100 queries per scene. Apartment scenes were selected due to the high diversity of objects from various room types within a single scene --- a better representation of environmental complexity common to robot deployments. For each scene we use between 100 and 1000 observations uniformly down-selected from the full set of images. Positive queries target objects present in the ScanNet label set for each scene. Negative queries were constructed using a combination of hand-authored labels and terms sampled from the ScanNet label set. The sampled labels were drawn from objects that appear elsewhere in the dataset but are absent from the specific scene. The labels used for the evaluation have a 70\% positive and 30\% negative split. Further evaluations using generic and non-generic queries were also conducted. Generic queries are short noun phrases (i.e. microwave, shoes), while non-generic queries include additional specifiers such as color ("blue cup") or immediate surroundings ("backpack by the couch").

\subsection{Experimental Setup}

Standard F1 and recall scores do not directly penalize systems that always return a result. The evaluation, therefore, focuses on the harmonic mean of recall and specificity, $F_{1s}$, an F1-style score that replaces precision with specificity, denoted with an $s$.

\begin{equation}
F_{1s} = \frac{2 \cdot \text{specificity} \cdot \text{recall}}{\text{specificity} + \text{recall}}
\end{equation}

This metric is used because it jointly rewards correctly retrieving positive instances and correctly abstaining negative ones. Metrics based solely on precision and recall over the positive class do not capture the latter. We compute specificity using the standard definition, meaning systems that never abstain are not rewarded for coverage. Any positive query which the system fails, either by abstaining or returning an incorrect result, is treated as a miss. The harmonic mean thus penalizes systems that abstain infrequently. This is practically motivated: a robot must navigate to search, and false confidence wastes navigation effort.

Two retrieval methods, \textit{CLIP} and \textit{Caption}, are combined with four selection methods (Table~\ref{tab:combinations}: \textit{Top-1}, \textit{SAM3}, \textit{LLM}, and \textit{LLM+SAM3}) for a total of 8 combinations. A 9$^{th}$ configuration, \textit{SAM3-all}, exhaustively runs SAM3 on every image in the observation set without a retrieval step. \textit{SAM3-all} serves as a reasonable baseline, requiring substantially more calls to SAM3 to deliver a functional utility for object retrieval. Note that by definition \textit{Top-1} configurations have no ability to abstain and therefore will have a specificity and $F_{1s}$ score of $0$. These methods are instead used as a baseline for secondary metrics such as precision.

\begin{table}[h]
\caption{All evaluated method configurations with key properties.}
\label{tab:combinations}
\begin{center}
\small
\begin{tabular}{@{}lccc@{}}
\toprule
\textbf{Method} & \textbf{Retrieval} & \textbf{Reject?} & \textbf{Visual?} \\
\midrule
CLIP/Top-1       & CLIP & No  & No \\
CLIP/LLM         & CLIP & Yes & No\\
CLIP/SAM3        & CLIP & Yes & Yes \\
CLIP/LLM+SAM3    & CLIP & Yes & Yes \\
\midrule
Caption/Top-1    & Caption & No  & No \\
Caption/LLM      & Caption & Yes & No \\
Caption/SAM3     & Caption & Yes & Yes \\
Caption/LLM+SAM3 & Caption & Yes & Yes \\
\midrule
SAM3-all         & None & Yes & Yes \\
\bottomrule
\end{tabular}
\end{center}
\end{table}

%%%%%%%%%%%%%%%%%%%%%%%%%%%%%%%%%%%%%%%%%%%%%%%%%%%%%%%%%%%%%%%%%%%%%%%%%%%%%%%%
\subsection{Results}

\begin{figure}[t]
    \centering
    \includegraphics[width=0.95\columnwidth]{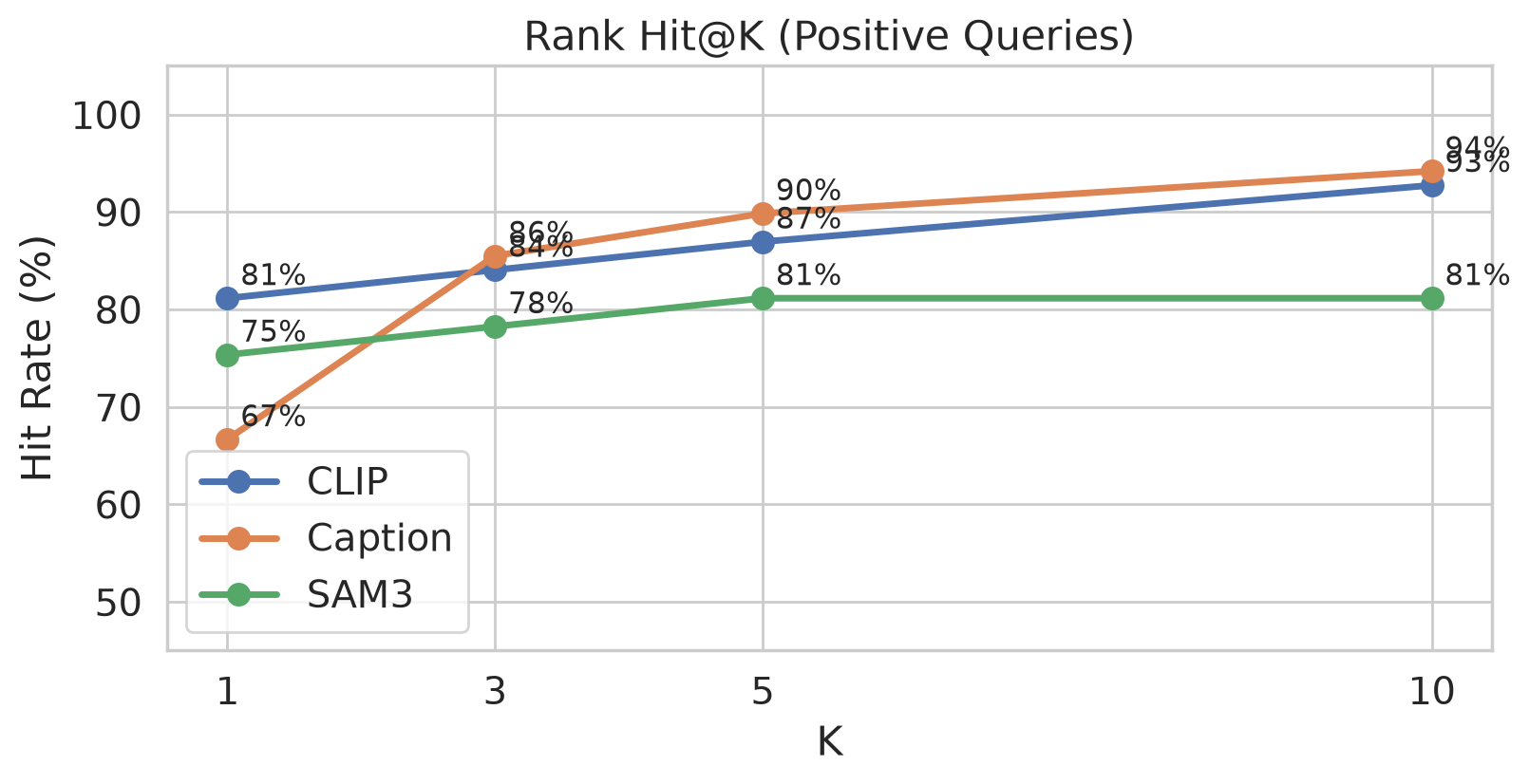}
    \caption{Retrieval Hit@K for CLIP, LLM-orchestrated caption search, and the SAM3 exhaustive baseline (ranked by detection confidence). CLIP leads at $K{=}1$ but caption search overtakes at $K{\geq}3$, suggesting broader coverage of relevant observations. All three curves exhibit an elbow at $K{=}5$. SAM3 has no gain from $K{=}5$ to $K{=}10$ and the retrieval methods add only $3$ to $6$ points beyond it, motivating our choice of $K{=}5$ as the operating point.}

    \label{fig:rank_hit}
\end{figure}

\begin{table*}[t]
\caption{Aggregate results across all 8 ScanNet scenes (averaged over 100 queries per scene). Best score in each column is \textbf{bold}.}
\label{tab:results}
\begin{center}
\small
\begin{tabular}{@{}lccccc@{}}
\toprule
\textbf{Method} & \textbf{$F_{1s}$} & \textbf{$F_1$} & \textbf{Specificity} & \textbf{Precision} & \textbf{Recall} \\
\midrule
CLIP/Top-1          &  0.0\%            & 66.3\%          &  0.0\%               & 56.0\%             & 81.2\% \\
CLIP/SAM3           & 68.4\%            & 69.0\%          & 83.3\%               & 85.1\%             & 58.0\% \\
CLIP/LLM            & 15.6\%            & 64.6\%          &  8.7\%               & 55.8\%             & 76.8\% \\
CLIP/LLM+SAM3       & 76.2\%            & 79.1\%          & 78.6\%               & 85.0\%             & 73.9\% \\
\midrule
Caption/Top-1       &  0.0\%            & 54.4\%          &  0.0\%               & 46.0\%             & 66.7\% \\
Caption/SAM3        & 75.5\%            & 75.9\%          & \textbf{92.5\%}      & \textbf{93.6\%}    & 63.8\% \\
Caption/LLM         & 15.3\%            & 64.2\%          &  8.5\%               & 55.2\%             & 76.8\% \\
Caption/LLM+SAM3    & \textbf{77.9\%}   & \textbf{81.8\%} & 77.5\%               & 85.7\%             & \textbf{78.3\%} \\
\midrule
SAM3-all            & 69.9\%            & 76.3\%          & 67.6\%               & 80.6\%             & 72.5\% \\
\bottomrule
\end{tabular}
\end{center}
\end{table*}

Methods without visual verification produce false positives on nearly all negative queries (Table~\ref{tab:results}). Using an LLM with previously generated captions yields a specificity of only 9\% regardless of the retrieval method. While using larger LLMs may improve this result, the gap is clear. Visual verification increases specificity to 68\% with the SAM3-all exhaustive search, and even further to 78\% when a retrieval stage is introduced. Caption based retrieval combined with SAM3 verification achieves a specificity of 92.5\%.

Visual verification reduces false positives on negative queries while still maintaining high overall performance on positive queries. All methods using visual verification have similar $F_{1s}$ scores ranging between 68\% (CLIP/SAM3) and 78\% (Caption/LLM+SAM3) while methods without are 52pp lower in the best case.

Fast retrieval combined with verified selection not only maintains performance comparable to the exhaustive search \textit{SAM3-all} baseline, but because \textit{SAM3-all} runs the detector over every observation, it has significantly more opportunities to fire a false positive on a negative query. The benefit therefore is improved reliability rather than a raw performance gain.

\begin{figure}[t]
    \centering
    \includegraphics[width=0.95\columnwidth]{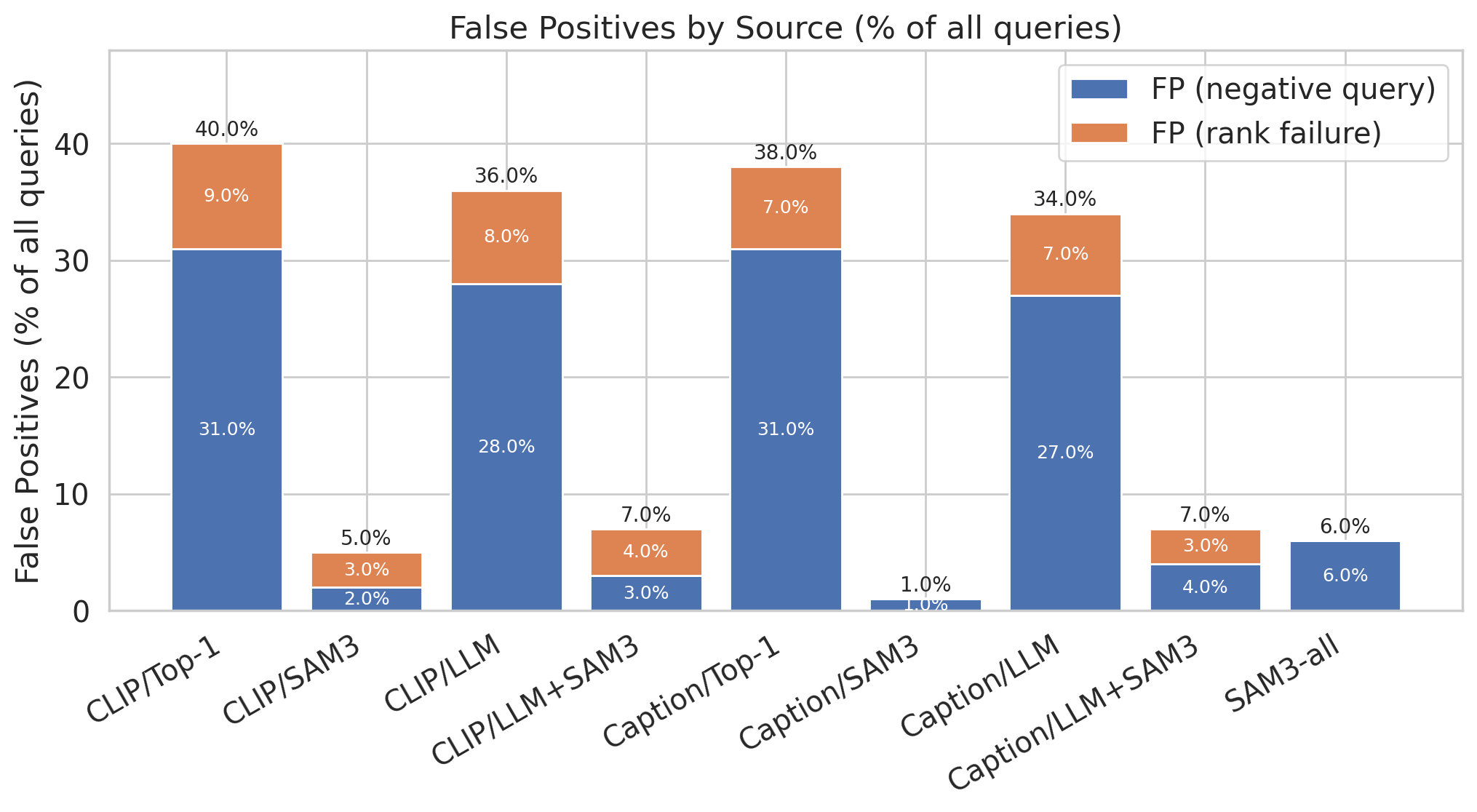}
    \caption{False positive analysis. Methods without visual verification (Top-1, LLM) produce false positives on nearly all negative queries. SAM3 verification (SAM3, LLM+SAM3) dramatically reduces false positives; the no-SAM3 methods have no visual verification step and therefore cannot reject semantically plausible but absent objects.}
    \label{fig:fp_analysis}
\end{figure}

Without any verification, all negative queries produce false positives (Fig.~\ref{fig:fp_analysis}). LLM reasoning alone reduces this only to roughly 91\%, confirming that text-based reasoning without visual confirmation remains insufficient. Adding visual verification via SAM3 drops false positives dramatically: to roughly 22\% with LLM+SAM3 and 8\% with SAM3 alone (Caption).

While CLIP shows a 16pp advantage over LLM captions for Hit@1 (Fig.~\ref{fig:rank_hit}), this gap reverses, but only by 2pp at $K=3$ and 3pp at $K=5$. Since $K=5$ is used for system evaluation, this small gap has an insignificant impact on overall performance.

CLIP's small increase of 6pp between $K=1$ and $K=5$ is likely due to its tendency to return temporally clustered images (consecutive frames of the same view), while Caption-retrieved candidates are more diverse across the scene, contributing to its increase of 25pp as $K$ increases.

\subsection{Robustness Across Query Types}

\begin{figure}[t]
    \centering
    \includegraphics[width=0.95\columnwidth]{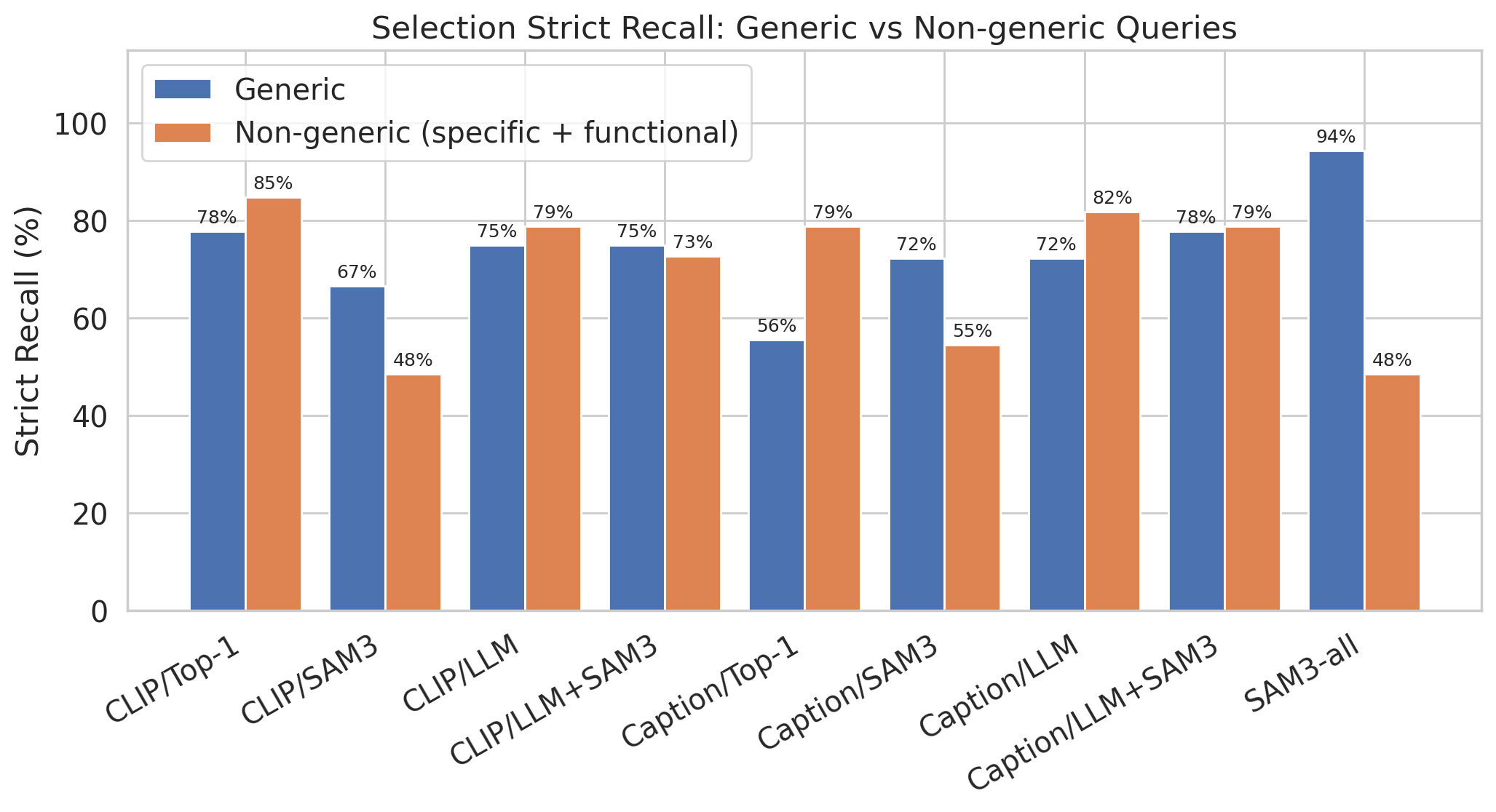}
    \caption{Recall on generic (short noun phrases) vs.\ non-generic (descriptive) queries by method. LLM-based selection methods maintain recall across query types; vision-only (SAM3) selection degrades significantly on non-generic queries.}
    \label{fig:generic_nongeneric}
\end{figure}

\textit{SAM3-all} has excellent recall of 92\% on generic, short noun phrases but drops sharply as the query complexity increases (Fig.~\ref{fig:generic_nongeneric}). We see a similar drop in the CLIP/SAM3 ($-$19pp) and Caption/SAM3 ($-$17pp) methods confirming the failure is in the SAM3 selection method. This is an expected limitation of SAM3 as open-vocabulary detection models are not designed to reason over queries, only identify them. Methods using LLMs are more robust and have similar recall for both query types. A robot using an LLM as its reasoning layer can pass the full range of query types into this system.

\section{REAL-ROBOT VALIDATION}

\subsection{Platform and Environment}

The architecture was validated on a small differential drive robot equipped with a 1080p wide-angle camera mounted approximately 1m above the ground, positioned to capture surfaces at countertop height, and a 2D lidar for localization. The differential drive robot base is a convenience stand-in for mobile manipulators and humanoids as the capability depends only on posed RGB images and transfers directly to other platforms. All inference ran locally using an NVIDIA 1080 Ti and 32 GB RAM, a modest, practically deployable configuration. SLAM pose was provided by SLAM Toolbox~\cite{macenski2021slam} via the ROS2 Nav2 stack, and observations were stored as image frames paired with pose and timestamp.

The environment is an on-campus hallway setting comprised of three rooms connected by a primary hallway and a fourth room connected by a secondary hallway, totaling approximately $150\,\text{m}^{2}$. The robot explored this previously unseen environment first to generate a map, then a second time to generate observations with poses. The second pass covered the space for approximately 15 minutes, yielding approximately 450 observations collected at $0.5\,\text{Hz}$.

The \textbf{CLIP/SAM3} method was chosen for initial live on-board deployment as it is the fastest configuration with visual verification and requires no VLM captions. All nine configurations were then evaluated post-hoc on the same observation set, enabling a direct comparison with the ScanNet results.

36 queries (21 positive, 15 negative) were constructed following a similar positive/negative split used for the ScanNet evaluation. In addition to common objects, several highly descriptive queries targeting objects unique to this environment were included.

\subsection{Results}

\begin{figure}[t]
    \centering
    \includegraphics[width=0.95\columnwidth]{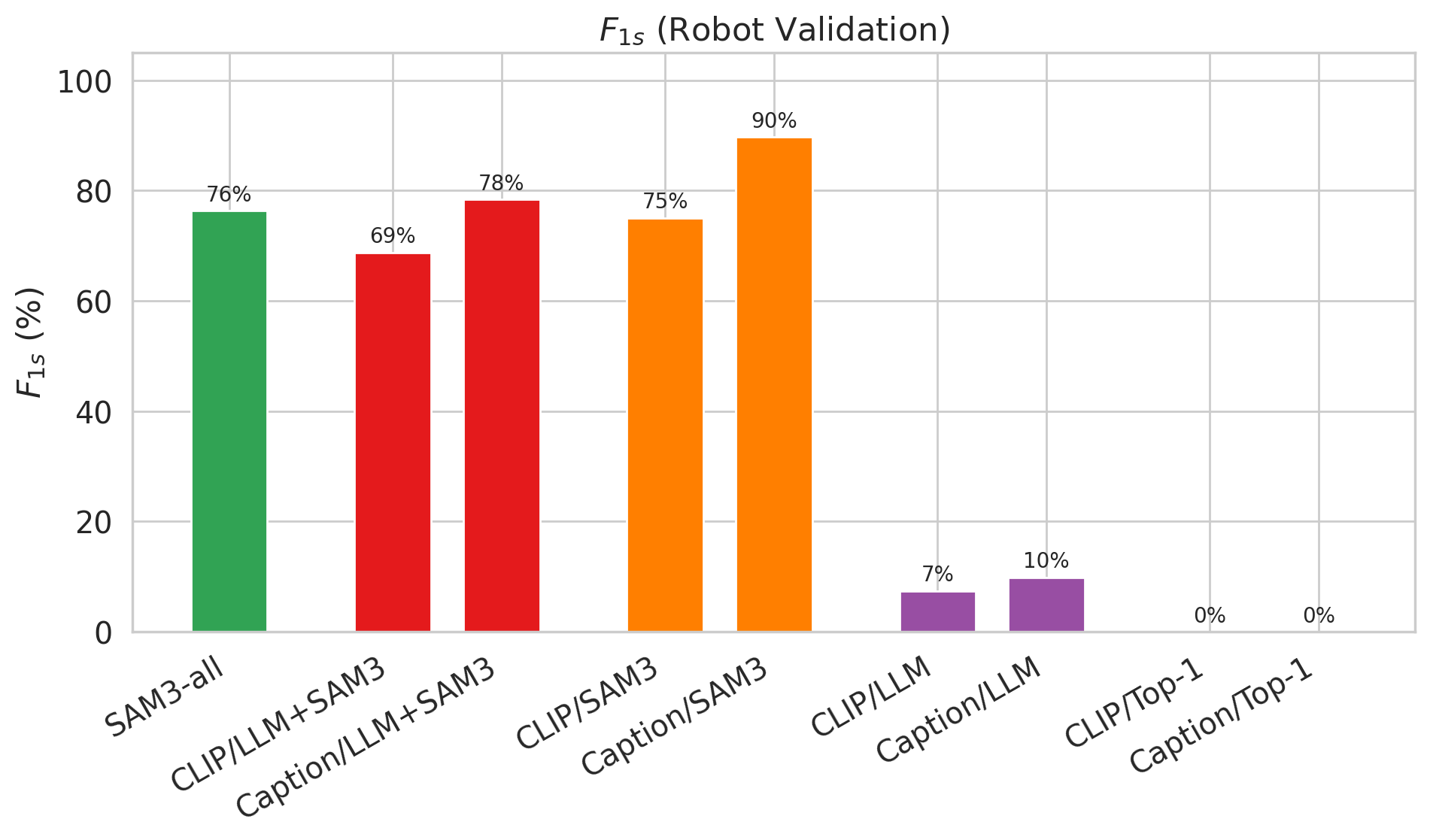}
    \caption{$F_{1s}$ scores on the real-robot scene (36 queries, 21 positive, 15 negative). Caption/SAM3 achieves the highest $F_{1s}$ of 89.7\%, exceeding its ScanNet counterpart and the exhaustive SAM3-all baseline.}
    \label{fig:robot_eval}
\end{figure}

\begin{table}[h]
\caption{Results on the real-robot scene (36 queries: 21 positive, 15 negative).}
\label{tab:robot_results}
\begin{center}
\small
\begin{tabular}{@{}lccc@{}}
\toprule
\textbf{Method} & \textbf{$F_{1s}$} & \textbf{Specificity} & \textbf{Recall} \\
\midrule
CLIP/Top-1          &  0.0\% &  0.0\% & 47.6\% \\
CLIP/SAM3           & 75.0\% & 95.2\% & 61.9\% \\
CLIP/LLM            &  7.4\% &  4.0\% & 52.4\% \\
CLIP/LLM+SAM3       & 68.8\% & 86.4\% & 57.1\% \\
\midrule
Caption/Top-1       &  0.0\% &  0.0\% & 66.7\% \\
Caption/SAM3        & \textbf{89.7\%} & 94.1\% & \textbf{85.7\%} \\
Caption/LLM         &  9.9\% &  5.3\% & 81.0\% \\
Caption/LLM+SAM3    & 78.4\% & 72.2\% & 85.7\% \\
\midrule
SAM3-all            & 76.3\% & 68.8\% & 85.7\% \\
\bottomrule
\end{tabular}
\end{center}
\end{table}

The relative ordering of methods on the robot dataset (Table~\ref{tab:robot_results}, Fig.~\ref{fig:robot_eval}) is consistent with the ScanNet results. Caption-based retrieval again outperforms CLIP across all selection methods at the cost of additional compute. Caption/SAM3 achieves the highest $F_{1s}$ of 89.7\% with a specificity of 94.1\% and only 1 false positive on 15 negative queries. The performance of the retrieval-gated configurations meets or exceeds the exhaustive SAM3-all baseline (76.3\%) while requiring SAM3 inference on only $K{=}5$ candidates rather than the full observation set.

\subsection{Qualitative analysis}

\begin{figure}[t]
    \centering
    \includegraphics[width=0.95\columnwidth]{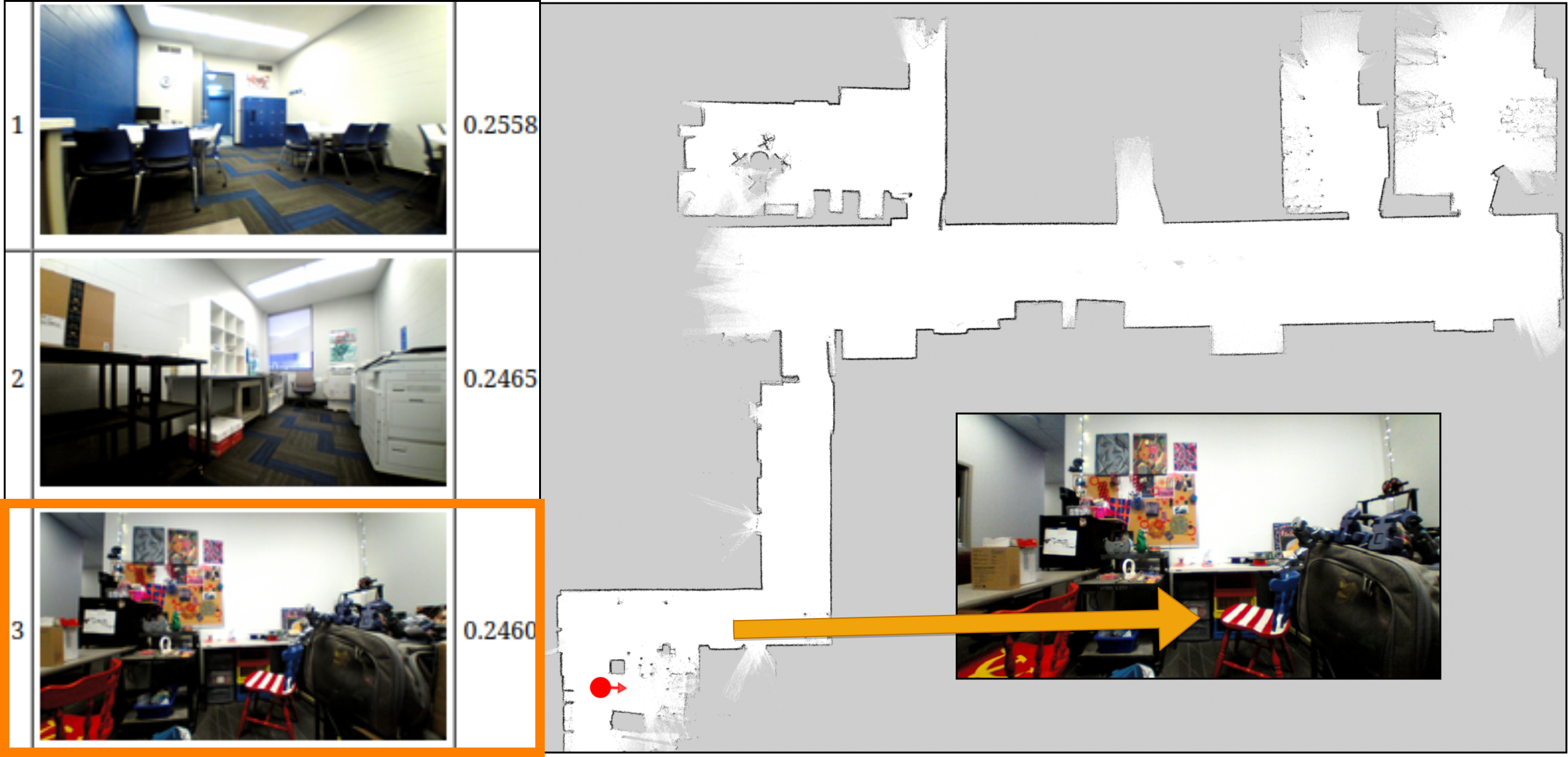}
    \caption{An American-flag-themed chair requiring joint color, pattern, and object-class reasoning. Retrieved by CLIP as the 3rd highest scoring result after 2 false positives (left). SAM3 verified the 3rd result and the correct target location was selected (right).}
    \label{fig:red-white-blue}
\end{figure}

\begin{figure}[t]
    \centering
    \includegraphics[width=0.95\columnwidth]{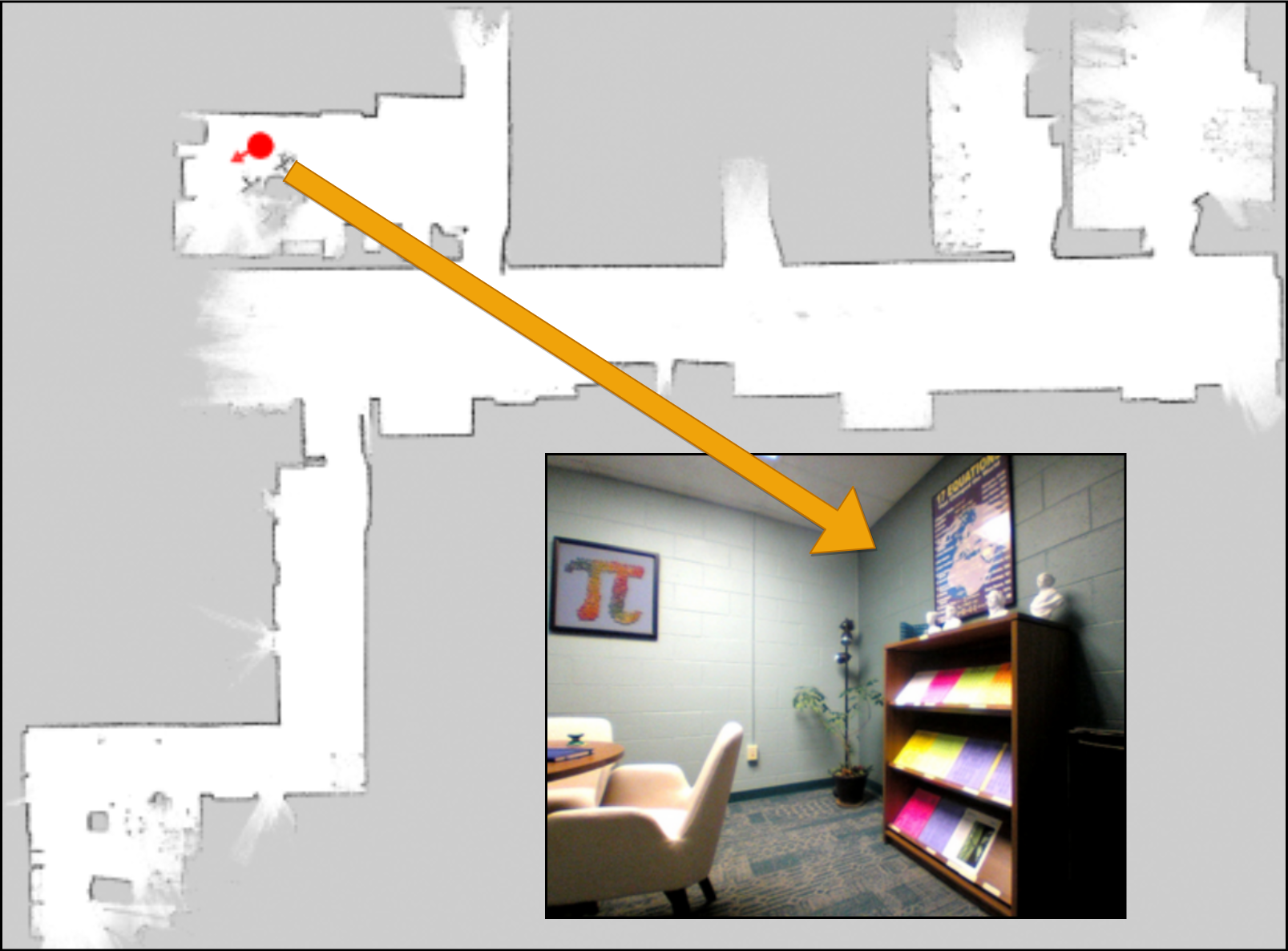}
    \caption{A "17 Equations" poster on the wall in a department office and the robot's selected target pose based on the query \textit{"17 Equations Poster"}.}
    \label{fig:poster}
\end{figure}

Several queries used for the live verification demonstrated the importance of the two stage architecture. Figure~\ref{fig:red-white-blue} shows the result of the query \textit{"Red, white, and blue painted chair"}, a unique American-flag themed chair located in a student workshop. CLIP identified the chair, but only as the third highest ranked result. The first and second highest results were false positives. SAM3 verified the chair in the 3rd result and rejected it in the top 2 results therefore correctly selecting the relevant observation and target pose.

For the "17 Equations" poster query, the CLIP retrieval stage retrieved incorrect observations even at $K{=}20$, while caption retrieval surfaced the correct one at $K{=}1$, which SAM3 then confirmed. The system correctly identified a target with the poster in view (Fig.~\ref{fig:poster}).

Several plausible but absent negative queries were also tested on the live system. The system correctly rejects queries like "vending machine", "label maker", "dry erase markers". In many cases, absent objects like "scissors" and "label maker" were selected by methods without visual verification but suppressed when visual verification was applied.

\subsection{Query Latency}

Table~\ref{tab:timing} reports per-query latency on the on-board hardware. An initial pre-processing step requires processing observations with CLIP (visual retrieval) or a VLM (caption-based retrieval). CLIP runs at 2-5 Hz, with the VLM significantly slower at 0.1 Hz. For VLM-based retrieval, observations were recorded at a reduced rate matched to this throughput, resulting in fewer total observations while still running in real time.

After pre-processing, however, retrieval (both visual and caption-based) is practically instant compared to the verified selection phase. SAM3 processes each candidate in approximately 0.4\,s, yielding approximately 2\,s for $K{=}5$. LLM inference adds approximately 8\,s per query. End-to-end latency for both stages is approximately 2\,s for SAM3-only selection and approximately 10\,s for LLM+SAM3. For live operation the robot uses CLIP/SAM3, which fits within the 2\,s budget without any pre-computation. 

These 2-10\,s latency solutions from our 2-stage architecture compare with the exhaustive SAM3 approach that takes 3 minutes per query on this dataset of 450 observations.

\begin{table}[h]
\caption{Per-query latency on on-board hardware (GTX 1080\,Ti).}
\label{tab:timing}
\begin{center}
\small
\begin{tabular}{@{}lc@{}}
\toprule
\textbf{Stage} & \textbf{Time} \\
\midrule
Retrieval (CLIP or caption)     & $<0.1$\,s \\
SAM3 inference ($K{=}5$)        & ${\approx}2$\,s \\
LLM reasoning (Gemma 3 12B)     & ${\approx}8$\,s \\
\midrule
End-to-end: SAM3-exhaustive     & ${\approx}3$\,min \\
End-to-end: CLIP+SAM3           & ${\approx}2$\,s \\
End-to-end: LLM+SAM3            & ${\approx}10$\,s \\\bottomrule
\end{tabular}
\end{center}
\end{table}

%%%%%%%%%%%%%%%%%%%%%%%%%%%%%%%%%%%%%%%%%%%%%%%%%%%%%%%%%%%%%%%%%%%%%%%%%%%%%%%%
\section{DISCUSSION}

Rather than introducing a new detector or embedding model, the contribution is the decoupled retrieval/verification pattern itself. This two-stage approach is a natural fit for LLM-orchestrated robot pipelines: since the retrieval stage accepts arbitrary text or visual embeddings and the selection stage provides grounded verification, the architecture can serve as a drop-in perception module in existing LLM-orchestrated systems. Because the selection stage treats the underlying detector as a swappable component, future improvements in open-vocabulary presence detection improve reliability of the full system directly.

Because the pipeline only depends on posed RGB observations, the same perception capability transfers directly to mobile manipulators and humanoid assistants performing tasks such as object retrieval in home and eldercare environments. The architecture also represents a practical middle ground between flat image retrieval and full 3D representations. 

The results from this work are evaluated with a single LLM (Gemma~3 12B); larger or more capable models may improve LLM-based selection, particularly on negative queries where text-only reasoning currently falls short, and locally run models are advancing rapidly. Our current benchmark is focused on ScanNet indoor apartment scenes, chosen as the most diverse individual scenes within the ScanNet scene types, but these are constrained environments with a well-defined object vocabulary; the most compelling near-term extension is evaluation across more diverse environments and longer observation sequences that stress-test retrieval coverage. A second direction is dynamic scene reasoning: given observation sets collected at $t_1$ and $t_2$, the system could detect which objects moved, appeared, or were removed, building directly on the existing retrieval and verification primitives. Third, a general spatiotemporal query interface (i.e.\ ``which rooms did you visit first?'', ``where should I put this book based on where you've seen similar books?'') can benefit from a system capable of confidently identifying or abstaining on queries.

\section{CONCLUSION}

We presented a two-stage architecture that decouples fast candidate retrieval from verified selection, enabling both high recall and reliable absence detection from sequential RGB observations with pose. Adding a visual verification step to caption-based retrieval matches or exceeds exhaustive search reliability while requiring an order of magnitude fewer detector calls, and we reproduce these ScanNet scores on a physical robot operating entirely on-board with a single consumer GPU running in real time. A robot acting on long-horizon memory must be able to determine, reliably, when something is or is not there. Decoupling fast retrieval from grounded verification shows that this reliability does not need to come at the cost of speed, making it a practical building block for open-ended, long-horizon tasks robots are increasingly asked to perform.

%%%%%%%%%%%%%%%%%%%%%%%%%%%%%%%%%%%%%%%%%%%%%%%%%%%%%%%%%%%%%%%%%%%%%%%%%%%%%%%%

\bibliographystyle{IEEEtran}
\bibliography{references}

\end{document}